\documentclass[12pt]{article}

\usepackage[margin=1in]{geometry}
\usepackage{xcolor}

\usepackage[margin=1in]{geometry}

\usepackage{graphicx}

\usepackage{amsmath, amssymb, amsfonts}

\usepackage[T1]{fontenc}
\usepackage{lmodern}
\usepackage{microtype}

\usepackage[hidelinks]{hyperref}
\hypersetup{
  pdftitle={LatentVerse: A Framework for Understanding Shared and Modality-Specific Information in Multimodal Latent Representations},
  pdfauthor={Majd Alafrange, Samuel Friedman, John Kitonyo, Sana Tonekaboni, Mahnaz Maddah}
}

\usepackage{caption}
\usepackage{subcaption}

\usepackage{enumitem}

\usepackage{booktabs}
\usepackage{xcolor}
\usepackage[numbers, square, sort]{natbib}

\begin{document}

\begin{flushleft}

{\LARGE \bfseries LatentVerse: A Framework for Understanding Shared
and Modality-Specific Information in Multimodal
Latent Representations\par}

\vspace{1em}

Majd Alafrange$^{1,2,3}$,
Samuel Friedman$^{1}$,
John Kitonyo$^{1}$,
Sana Tonekaboni$^{1,2,3, \dag}$,
Mahnaz Maddah$^{1,*, \dag}$

\vspace{0.8em}

{\small
$^{1}$Machine Learning for Health (ML4H), Broad Institute of MIT and Harvard, Cambridge, Massachusetts, USA\\
$^{2}$Massachusetts Institute of Technology, Cambridge, Massachusetts, USA\\
$^{3}$The Schmidt Center, Broad Institute of MIT and Harvard, Cambridge, Massachusetts, USA\\
$^{\dag}$Joint supervision\\
$^{*}$Corresponding author
}

\vspace{1.2em}

\fcolorbox{black!20}{gray!8}{
\begin{minipage}{0.96\textwidth}
\textbf{Corresponding Author:}\\
Mahnaz Maddah, PhD\\
Senior Director, Machine Learning for Health\\
Broad Institute of MIT and Harvard\\
415 Main Street, Cambridge, MA 02142\\
Email: \href{mailto:maddah@broadinstitute.org}{maddah@broadinstitute.org}\\
Phone: (617) 714-7000
\end{minipage}
}

\end{flushleft}

\vspace{1.5em}

\begin{abstract}
Latent embeddings have become a central data abstraction in modern machine learning, especially in biomedicine, where foundation models are increasingly used to encode multimodal data like clinical text, medical images, omics, and physiological signals. However, the utility and value of these representations depends on understanding their quality, structure, and the information they encode. Existing analysis workflows for evaluating representations remain fragmented across custom scripts, isolated metrics, and most importantly lack multimodal analysis, limiting accessibility and reproducibility. We present LatentVerse, a representation analysis resource that combines a web-based visual analytics platform for accessible, report-driven exploration with a command-line interface for scalable technical workflows. LatentVerse unifies diagnostics for various representation quality metrics and extends to multimodal settings by decomposing embeddings into shared and modality-specific components. We evaluate LatentVerse through controlled unimodal and multimodal simulations, discovery-oriented analyses on real biomedical embeddings, and a user study across diverse use cases. By supporting thorough and interpretable evaluation of latent spaces, LatentVerse makes foundation model representations more understandable in biomedical and data science applications.
\end{abstract}
\newpage

\section{Introduction}

Latent representations (embeddings) have become a standard data
abstraction in modern machine learning, in applications ranging
from natural language to vision \citep{bengio2013representation}. In particular, the
recent shift toward large-scale pretraining followed by task-specific adaptation has further increased their importance
\citep{brown2020language, alayrac2022flamingo}. Embeddings enable knowledge discovery \citep{johnson2026embeddings},
are reused across tasks \citep{rasmy2021med, beam2020clinical}, compared across models and checkpoints in zero-shot settings \citep{kedzierska2025zero}, and shared across teams or institutions when raw data cannot be easily accessed \citep{di2025embedding}. As a result,
understanding the quality of an embedding space has become an important part of modern machine learning workflows.

This need is especially pressing when downstream evaluation alone is insufficient \citep{you2021logme}. Practitioners often need to compare representations before expensive fine-tuning and assess whether embeddings remain reliable under limited labels, noisy inputs, or domain shift. These challenges are even greater in multimodal settings, which are becoming increasingly common across real-world machine learning applications. Unlike unimodal embeddings, multimodal representations may entangle information from various sources, yet there are few established tools for systematically evaluating this structure. Representation-centric diagnostics, therefore, provide a more direct way to evaluate latent spaces, using metrics such as clusterability, disentanglement, robustness, expressiveness, and predictability to characterize how information is organized, preserved, and exposed within both unimodal and multimodal embeddings.

Recent scientific machine learning resources have highlighted the growing value of accessible and reproducible analysis platforms, yet existing tools only partially address representation evaluation. General-purpose embedding viewers such as the TensorBoard Embedding Projector \citep{smilkov2016embedding}, Emblaze \citep{sivaraman2022emblaze}, WizMap \citep{wang2023wizmap}, and Embedding Atlas \citep{ren2025embedding} support interactive projection and neighborhood exploration, while domain-specific systems such as Latent Space Explorer and other scientific web applications provide guided analysis within particular modalities or application areas
\citep{kwon2023latent,cecconello2022latent,ospina2025spatialge}. In parallel, benchmarking and data platforms emphasize downstream task performance or large-scale data access rather than representation-centric
diagnosis \citep{jiang2025ukbmdrmf,czi2024czbenchmarks,abdulla2025cellxgene}.
Taken together, these tools establish the value of interactive exploration and accessible infrastructure, but none provides a unified,
model-agnostic workflow for systematic latent-space evaluation that spans unimodal and multimodal settings and can localize where information resides across modalities.

To address this gap, we present LatentVerse, a representation-analysis tool that builds on the original LatentVerse toolkit
\citep{turura2025latentverse}. LatentVerse integrates a broad family of representation diagnostics into a unified, model-agnostic framework and extends it to multimodal representations through
information decoupling \citep{wanginformation, tonekaboni2026multiloreft},
which separates shared from modality-specific latent information so that
each component can be analyzed independently. This makes it possible to reason directly about where useful information resides in multimodal
systems and how different representation components contribute to performance and interpretability. 

We evaluate LatentVerse along three axes: controlled
unimodal and multimodal simulations that test whether its diagnostics recover known structure; real biomedical embedding analyses that show its utility for discovery; and a user study
that assesses its practical value in analytical workflows. By integrating these capabilities into a single workflow, LatentVerse helps bridge the gap between representation learning research and applied practice.

\section{The LatentVerse Framework}
\label{sec:framework}

\begin{figure}[!h]
\centering
\includegraphics[width=\textwidth]{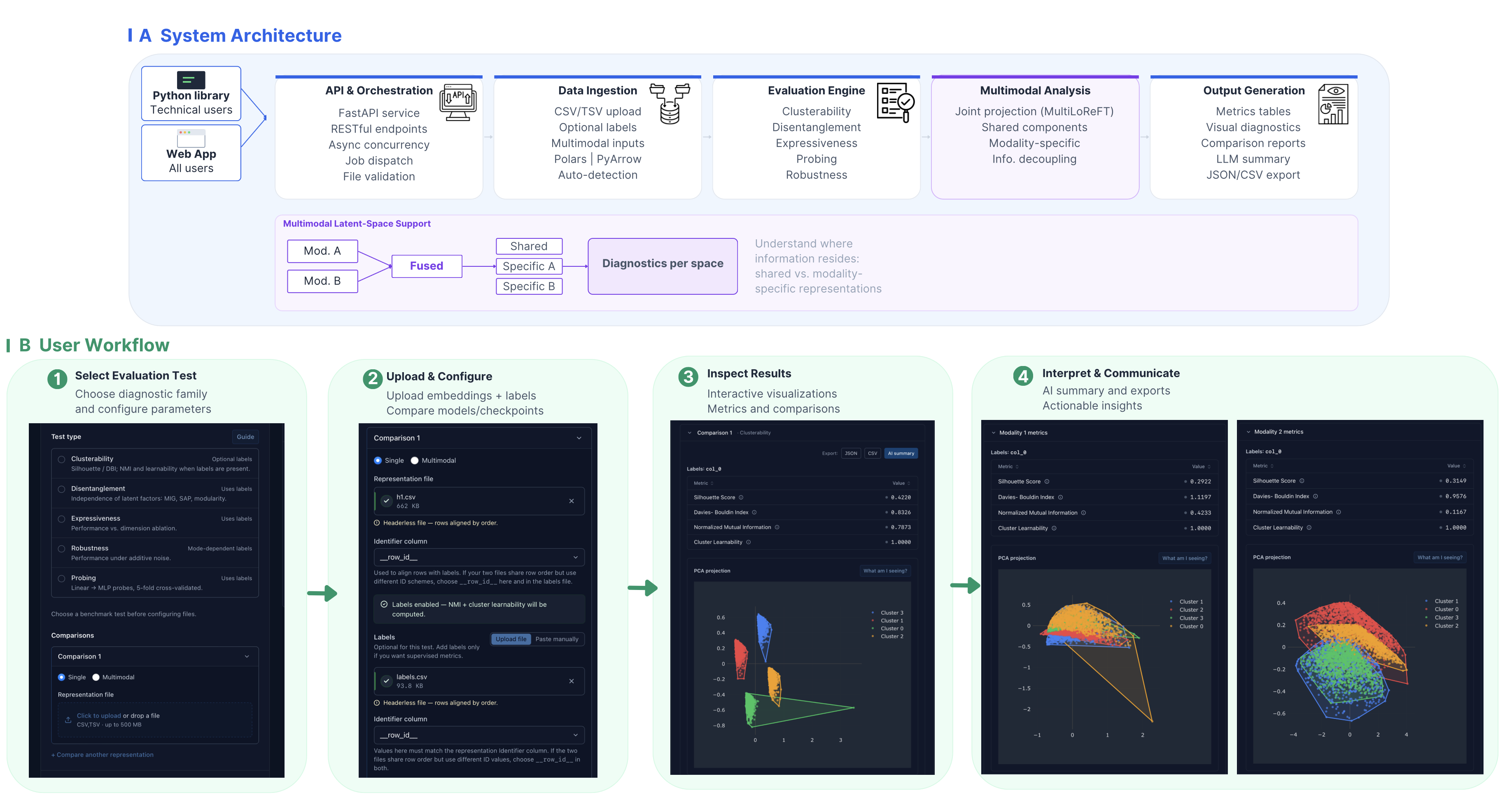} 
\caption{\textbf{Overview of the LatentVerse framework.}
A) System architecture showing the shared evaluation pipeline, from web
and Python-library access through data ingestion, representation
diagnostics, multimodal decomposition, and report generation.
B) User workflow, from selecting an evaluation test and uploading
embeddings to inspecting results and communicating findings through
exportable reports and AI-generated summaries.}
\label{fig:overview}
\end{figure}

LatentVerse is a model-agnostic framework composed of a web application and a command-line interface for evaluating latent representations. Users provide one or more representation matrices and optional labels, select a diagnostic test, and receive a structured report combining quantitative metrics, interactive visualizations, and a descriptive summary. The framework supports both unimodal evaluation and multimodal analysis described in Section~\ref{sec:multimodal}.
Figure~\ref{fig:overview} provides an overview of the system and user workflow. The high-level framework is described in the following sections, and mode detail on implementation, deployment, and interface engineering is presented in Appendix~\ref{app:system}.

\subsection{Representation quality dimensions}
\label{sec:quality}
LatentVerse evaluates five complementary properties of latent representations: clusterability, disentanglement, expressiveness, downstream predictability, and robustness. Together, these provide a multi-faceted view of representation quality that goes beyond downstream task accuracy alone. LatentVerse uses established metrics and tests for each evaluation category to provide a consistent workflow. Detailed description and definitions of metrics are provided in Appendix~\ref{app:tests}.

\paragraph{Clusterability.} The clusterability test evaluates whether a
representation forms compact and well-separated groups in latent space.
LatentVerse combines intrinsic metrics, including Silhouette Score \citep{rousseeuw1987silhouettes} and
Davies--Bouldin Index \citep{davies1979cluster}, with label-aware measures such as Normalized
Mutual Information (NMI) \citep{strehl2002cluster} and Cluster Learnability \citep{lu2023using}. The resulting report
helps users determine whether the embedding supports meaningful grouping
structure and whether that structure aligns with known labels.

\paragraph{Disentanglement.} The disentanglement test evaluates whether
distinct factors of variation are encoded in separate and interpretable
latent dimensions. LatentVerse reports complementary metrics including
the DCI framework \citep{eastwood2018framework}, Mutual Information Gap (MIG)
\citep{chen2018isolating}, Separated Attribute
Predictability (SAP) \citep{kumar2018variational}, and Total Correlation (TC)
\citep{watanabe1960information}. These outputs help
users distinguish representations that are cleanly factorized from those
in which relevant information is spread across many correlated dimensions.

\paragraph{Expressiveness.} The expressiveness test evaluates how
efficiently predictive information is distributed across the latent
dimensions. To measure this, LatentVerse performs dimension ablation by progressively removing the highest-variance dimensions and measuring the resulting drop in predictive performance, summarizing this behavior through Compactness and Intrinsic Dimension. This lets users assess whether useful signal is
concentrated, redundant, or diffusely distributed.

\paragraph{Downstream predictability.} The downstream predictability test
measures how much task-relevant information can be recovered from the
latent space. LatentVerse evaluates a family of supervised probes of
increasing complexity, from linear models to multilayer perceptrons, and
reports standard predictive metrics such as Accuracy, AUROC, macro-F1,
and $R^2$, depending on the task. This distinguishes representations that
are directly predictive, those that require more complex decoders, and
those that contain little usable signal for the target label.

\paragraph{Robustness.} The robustness test evaluates how stable a
representation remains under perturbations. LatentVerse injects Gaussian
noise into the latent space and tracks the resulting degradation in
clustering- or probing-based performance. The corresponding degradation
curves reveal whether an embedding is resilient or fragile under noise,
providing a practical view of representation reliability.

\subsection{A unified evaluation workflow}
\label{sec:workflow}
The framework is organized around a three-step workflow. First, the user
selects the type of representation they want to evaluate (unimodal or multimodal) and an evaluation test along with the test-specific options.
Second, they upload one or more representation files together with
optional labels and comparison settings. Third, LatentVerse executes the
requested analyses and renders the results as a report consisting of an
interactive visualization, a metrics table, exportable outputs, and an
optional descriptive summary. Crucially, the same
pipeline drives unimodal and multimodal analysis and is exposed
identically through the web application and a command-line interface, so
that analyses can move between accessible exploration and scalable,
scriptable workflows without changing the underlying diagnostics. A parity test suite verifies that the two interfaces produce numerically
identical results---agreement to a relative tolerance of $10^{-9}$ on the same inputs---so an analysis can be developed interactively and rerun from a script without change. Implementation details that support practical use on large embedding matrices, scalable ingestion, asynchronous job execution, and resumable uploads, are described in
Appendix~\ref{app:system}.

\subsection{Visual analytics and report generation}
\label{sec:visual}
The interface is designed to support both technical and less technical
users. Guided configuration, built-in documentation, example reports, and
contextual explanations lower the barrier to entry, while comparison
cards, advanced options, and export functionality support more detailed
analytical workflows. Each report couples three
coordinated views, a metrics table, a test-specific interactive
visualization, and a on-click natural-language summary, so that numerical
evidence, visual evidence, and interpretation are presented together
rather than as isolated artifacts. Across all tests, the goal is to make
the properties of a latent space easier to inspect, compare, and
interpret. The summary is produced by a server-side, template-constrained
query to a large language model; details of the visualization types and
the summary mechanism are given in Appendix~\ref{app:interface}.

\subsection{Multimodal analysis}
\label{sec:multimodal}

The most notable capability of LatentVerse is its
treatment of multimodal representations. Modern machine learning systems
increasingly integrate multimodal sources like text, images, audio, or
clinical measurements, and the resulting embeddings can be more informative than unimodal ones. They are also substantially harder to interpret because a single fused vector may entangle information that is shared across modalities with information that is specific to each modality.

Prior work in factorized multimodal representation learning has proposed
information decoupling as a way to improve interpretability by separating
shared factors that capture cross-modal agreement from modality-specific
factors that retain unique information from each input stream
\citep{wanginformation, tonekaboni2025fusion}. LatentVerse
integrated this capability into its workflow to separate the shared and modality-specific information in mutlimodal representation and applies
the same family of diagnostics used in the unimodal case not only to a fused multimodal embedding, but also to decomposed components.

To obtain this decomposition, LatentVerse leverages the MultiLoReFT
framework \citep{tonekaboni2026multiloreft}, a lightweight fine-tuning
approach for separating shared and private information in multimodal
representations. MultiLoReFT isolates information unique to each modality from information shared across modalities which enables representation evaluation on each component independently. The detailes of our experiments are reported in Appendix~\ref{app:multiloreft}, Table~\ref{tab:multiloreft}.

Decomposition changes the kinds of questions that can be
asked of a multimodal representation. Rather than asking only \emph{how
predictive is this embedding?}, a user can ask whether a clinically
relevant factor is encoded in the shared space or in one modality's
private space; whether predictive structure is genuinely cross-modal or
driven by a single modality; and whether shared and private components
differ in compactness or robustness. By evaluating each component
separately, LatentVerse makes these questions answerable from the data
rather than assumed. 

\begin{figure}[t]
\centering
\includegraphics[width=\textwidth]{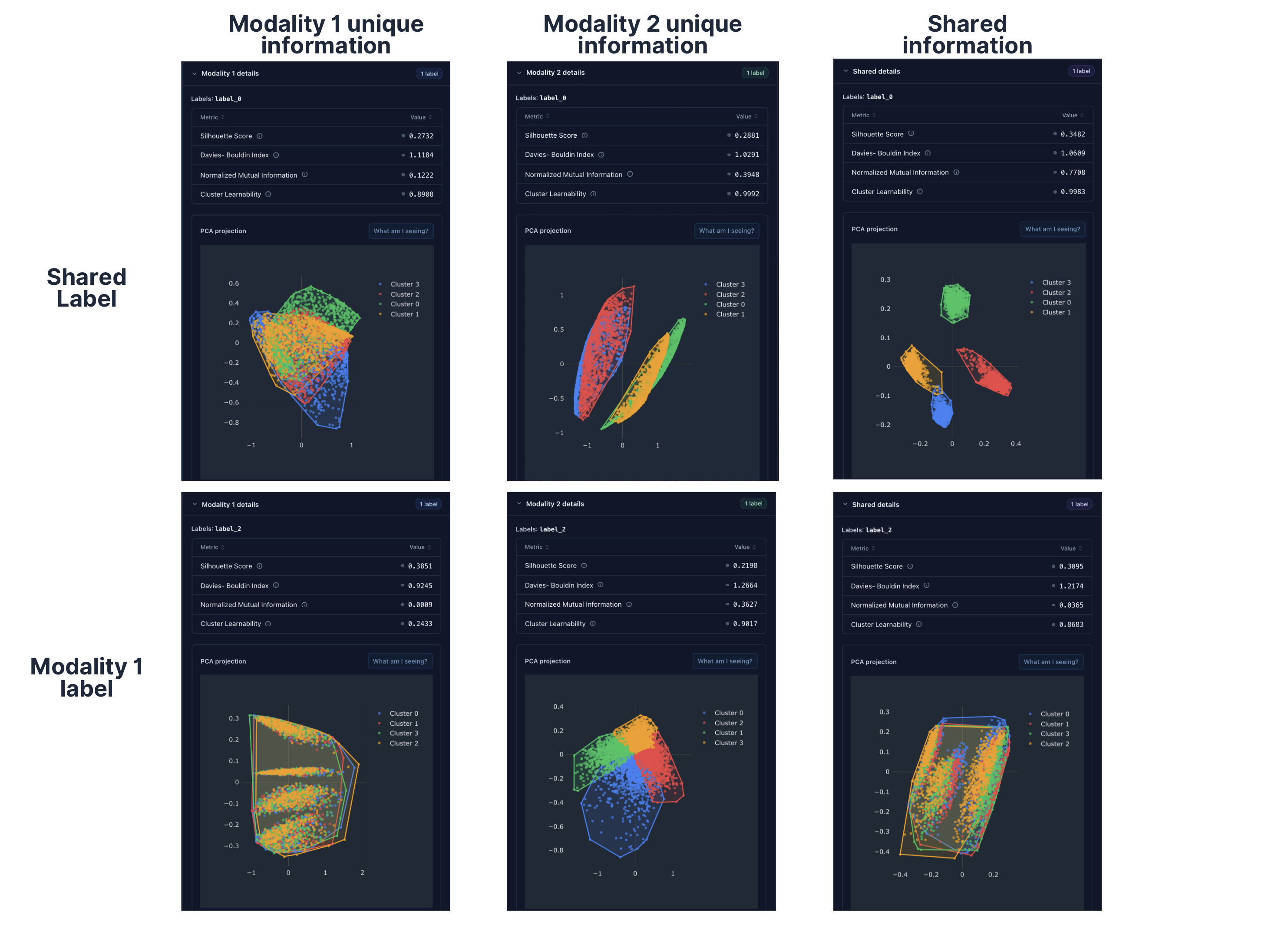}
\caption{\textbf{LatentVerse multimodal analysis.} LatentVerse decomposes
information from multiple representations into shared and
modality-specific components. By evaluating each component separately,
the platform helps users identify where shared and modality-specific
information resides in latent space.}
\label{fig:multimodal}
\end{figure}

Figure~\ref{fig:multimodal} shows how LatentVerse decomposes a multimodal representation into shared and modality-specific components and how the behavior of each component reveal variability in the underlying generative factors.
In this way, LatentVerse moves beyond treating a multimodal embedding as a
single fused object and instead turns it into an analyzable
representational object. This also makes the framework especially useful where researchers need to compare fusion strategies, validate whether modalities contribute complementary information, or
communicate why one multimodal model is preferable to another.

\section{Validation on Controlled Data}
\label{sec:validation}

We first validate LatentVerse on simulated
representations with well-defined and known structure. The purpose of these experiments is to confirm that the framework's integrated metrics
and visual reports recover the intended properties of a latent space, so
that further analysis with the goal of scientific discovery (Section~\ref{sec:biomedical}) rest on trusted
diagnostics. Full details of all unimodal and multimodal dataset generation are provided in Appendix~\ref{app:sim}.

\subsection{Unimodal simulations}
\label{sec:val-unimodal}
\begin{figure}[!h]
\centering
\includegraphics[width=\textwidth]{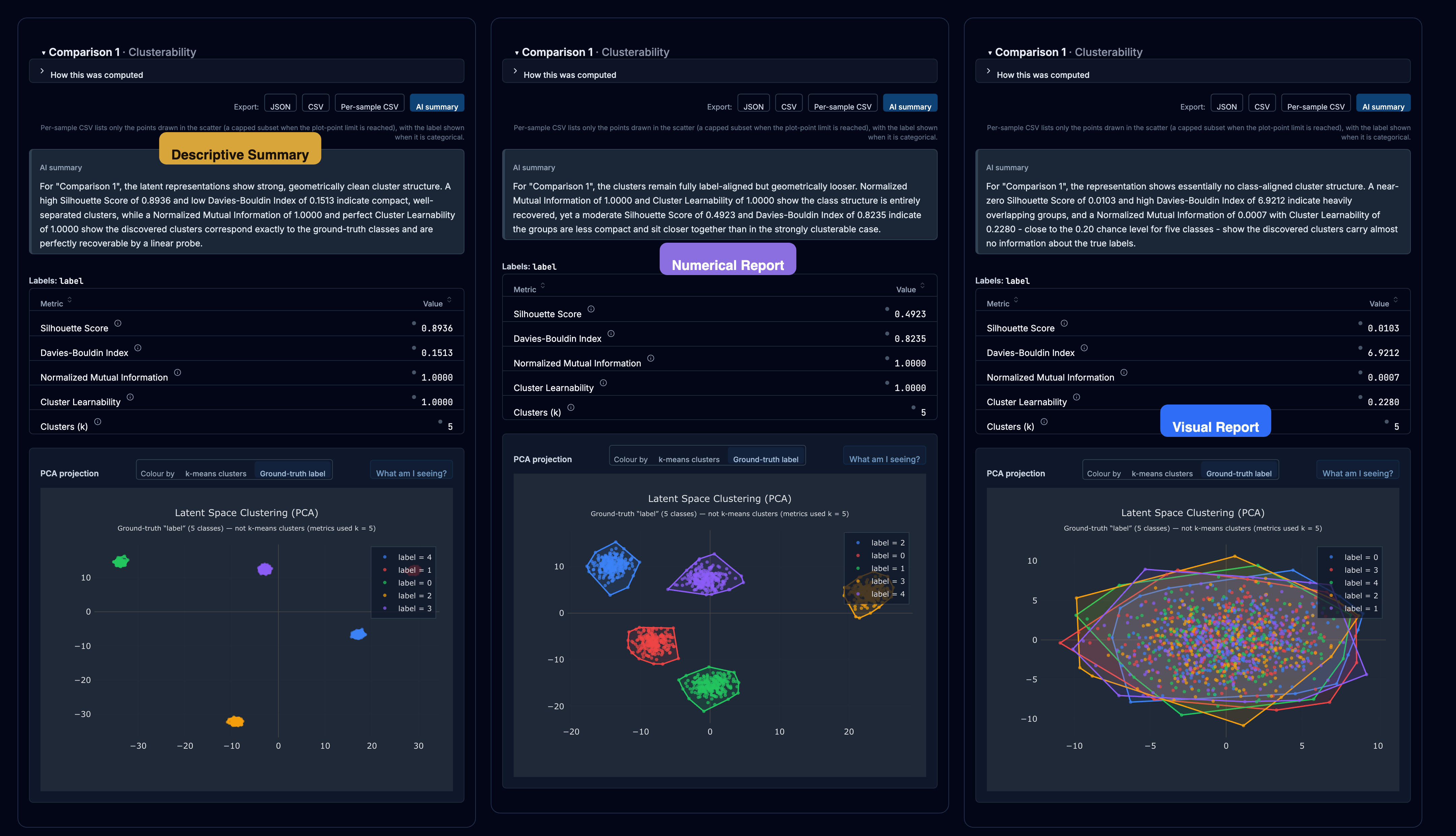}  
\caption{Validation on controlled simulated data. Example
LatentVerse reports for representations with varying clusterability. As
cluster structure degrades from left to right, the reported metrics
reflect weaker class-aligned grouping, the latent-space visualizations
transition from compact, well-separated clusters to diffuse and
overlapping point clouds, and the descriptive summary translates the
results into an interpretable explanation.}
\label{fig:clusterability}
\end{figure}

We generated dataset families that each target one representation
property (clusterability, predictability, disentanglement, and
expressiveness), with three variable levels per family ranging from
settings where the target property is strongly present to settings where it is weak or absent. Across these datasets, we show that LatentVerse provides evaluation reports that track the known structure, and helps identify such characteristics of the data. As a representative example, the clusterability datasets consists of representations that vary from clearly separated clusters to
overlapping groups to no class-aligned structure. 
Figure~\ref{fig:clusterability} shows LatentVerse reports for these representation groups. In the highly clusterable setting the Silhouette Score (0.8936) and NMI (1.0000) reflect clearly separated clusters, and Cluster Learnability (1.0000) confirms the class labels are almost perfectly recoverable from the representation; in the unstructured setting these fall to 0.0103, 0.0007, and 0.2280 respectively---the last near the 0.20 chance level for five classes---with the visual report showing the same progression from tight, well-separated groups to diffuse, overlapping point clouds. LatentVerse report also includes a descriptive summary report that analyses the evaluated metrics and provides an interpretable description of the representation behaviour according to the evaluation category. 
The predictability, disentanglement, and expressiveness
families behave similarly, distinguishing directly accessible from
decoder-dependent and uninformative labels, factorized from entangled
representations, and concentrated from redundant or sparse information
distributions, respectively. Because these results serve only to confirm
expected metric behavior, we report the full set of examples and figures
in Appendix~\ref{app:val-supplement} and summarize them here.

\subsection{Multimodal simulations}
\label{sec:val-multimodal}
We also evaluate LatentVerse in a multimodal setting designed to test
whether the framework can provide better insight into the multimodal datasets by decoupling shared from modality-specific
information. For this purpose, we generate a multimodel simulated dataset where the representations are generated from independent latent factors drawn from diverse distributions, with labels defined to be shared across modalities, specific to one modality, or conditionally dependent on interactions between modalities (See Appendix~\ref{app:sim-multimodal} for more detail on this dataset).
The experiment is intended to show that decomposition
yields a more informative view than evaluating only a fused embedding.
After decomposing the embedding into shared and modality-specific
components, the reports reveal that shared labels are concentrated in the shared latent space, while modality-specific labels are most strongly recovered in their corresponding private components (Figure \ref{fig:multimodal}). This separation is
reflected in both the reported metrics and the latent-space
visualizations, which makes it possible to inspect where different forms of
information reside. These controlled results establish
that the decomposition workflow recovers the ground-truth organization of information; this motivates its use on real embeddings where no such ground truth is available.

\section{Representation Analysis for Real-world Discovery}
\label{sec:biomedical}

\begin{figure}[t]
\centering
\includegraphics[width=\textwidth]{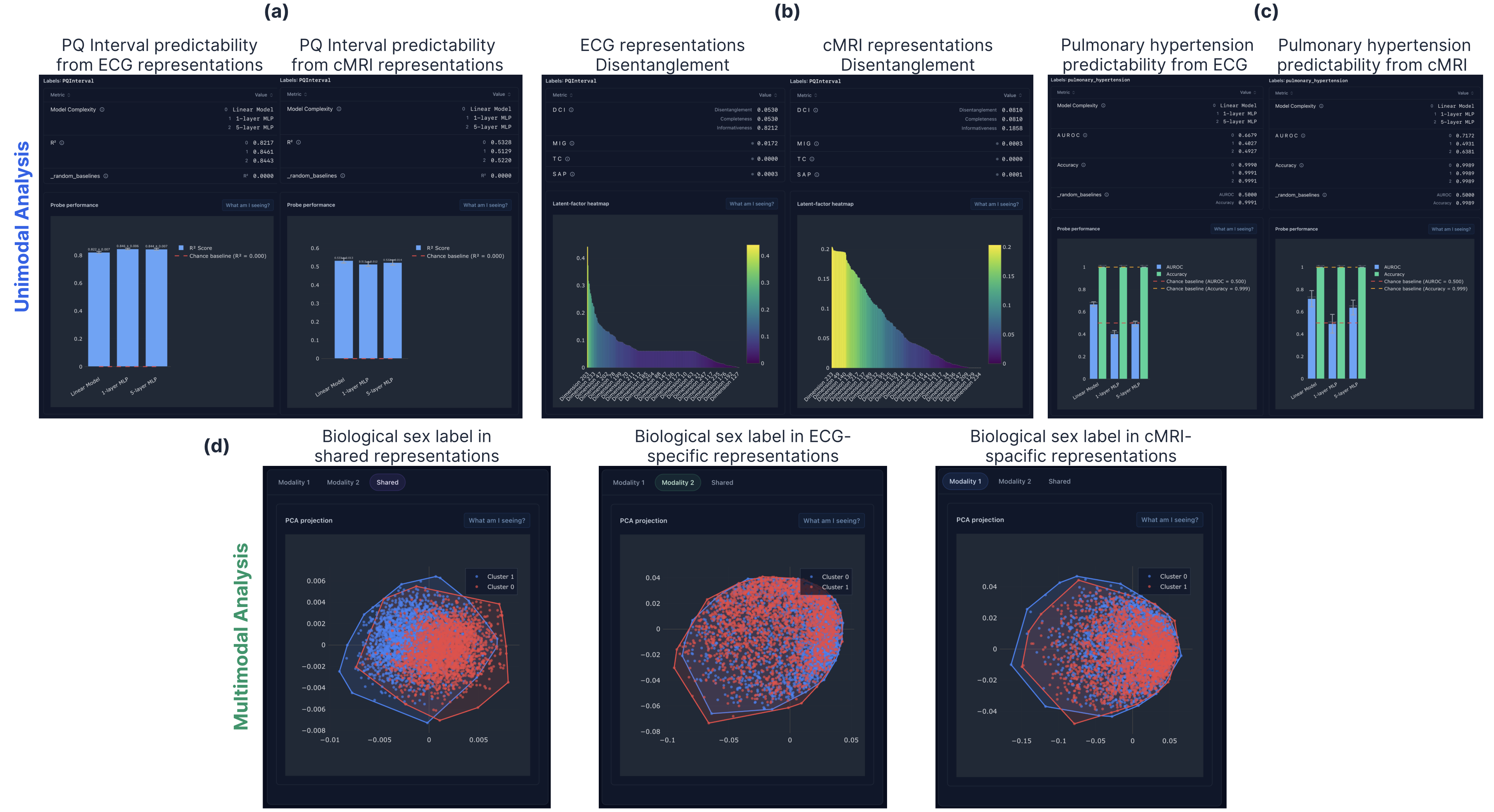} 
\caption{\textbf{LatentVerse analysis of real biomedical embeddings from
UK Biobank.} (a) PQ-interval predictability from ECG and cMRI
representations. (b) PQ-interval disentanglement in ECG and cMRI
representations. (c) Pulmonary-hypertension predictability from ECG and
cMRI representations. (d) Multimodal analysis of biological sex across
shared, ECG-specific, and cMRI-specific latent spaces, showing stronger
separation in the shared representation. These examples illustrate how
LatentVerse localizes clinically relevant information across modalities
and latent subspaces.}
\label{fig:real_biomed}
\end{figure}

We next apply LatentVerse to real biomedical embeddings to assess its value in better understanding and assessing real-world representations. We show that the diagnostic tests can reveal which modality carries the strongest signal for a given phenotype, whether that signal is shared across modalities or confined to one of them, how concentrated or diffuse it is within latent space, and how robust different embeddings are under analysis. These questions are highly relevant in biomedical machine learning, where pretrained representations are increasingly reused for downstream studies but are often treated as opaque features. To this end, we analyze pretrained embeddings from UK Biobank \citep{bycroft2018ukbiobank}, derived from electrocardiogram (ECG) signals and cardiac magnetic resonance imaging (cMRI) using pretrained encoder models. Figure~\ref{fig:real_biomed} summarizes the resulting analyses.

Figure~\ref{fig:real_biomed} illustrates several examples of this analysis. Figure~\ref{fig:real_biomed}.a shows differences in {predictive accessibility} across modalities. For the PQ interval, ECG representations achieve substantially higher predictive performance than cMRI representations, with strong $R^2$ values even for simple probes, indicating that this signal is directly accessible from ECG latent space. In contrast, cMRI representations are less predictive for the same target, suggesting that the relevant information is either weaker or encoded less directly. 

Figure~\ref{fig:real_biomed}.b provides insight into how information is distributed in latent dimensions through disentanglement analysis. For the same PQ-interval target, the ECG representation exhibits stronger evidence of concentrated, target-related structure, whereas the cMRI representation shows weaker disentanglement and a higher distribution of information across latent dimensions. The corresponding heatmaps make this distinction visually apparent by showing whether label-related signal is localized to a smaller subset of latent variables or spread more broadly across the embedding. 

Figure~\ref{fig:real_biomed}.c shows clinical predictability of different representations for pulmonary hypertension diagnosis. Both ECG and cMRI representations can be assessed through probing, but the reports also show why a single metric is insufficient: because the label is highly imbalanced, accuracy remains artificially high across probes, while AUROC is more informative about true predictive utility. In this example, cMRI representations appear to provide stronger discriminative signal than ECG representations, illustrating how the platform helps users identify which modality carries more clinically useful information.

The multimodal analysis further illustrates how LatentVerse can support discovery by localizing where specific biological information is encoded. Figure~\ref{fig:real_biomed}.d shows the distribution of biological sex labels across the shared, ECG-specific, and cMRI-specific representation components. In this case, the clearest separation appears in the shared latent space, while the modality-specific components show substantially weaker structure. This is consistent with the expectation that biological sex is reflected across both ECG and cMRI and therefore should be captured primarily in information shared between the two modalities. More broadly, this example shows how LatentVerse can help distinguish whether a clinically relevant variable is jointly encoded across modalities or driven mainly by one modality alone.

\section{User Study}
\label{sec:user-study}
To assess the practical utility of LatentVerse, we conducted a user study with researchers from diverse biomedical application domains who were already familiar with machine learning workflows. Rather than evaluating the system on a single task or dataset, the study was designed to understand how LatentVerse could support representation analysis in participants' own areas of work. Participants used the platform to examine embeddings drawn from a range of biomedical applications and then provided feedback on how its metrics, visual reports, and comparative views could support their existing analytical practice. The study therefore focused on the perceived usefulness, interpretability, and workflow fit of LatentVerse for real biomedical representation analysis, rather than only on generic usability. The full study protocol and survey instrument are provided in Appendix~\ref{app:survey}.

\subsection{Results}
\label{sec:user-results}

Three researchers took part: two working in genomics and proteomics and one in clinical machine learning, all of whom currently evaluate embeddings using
Python notebooks and custom visualisation scripts. Participants used the
platform as deployed in July 2026. Given this sample size, we report their
responses as formative design feedback rather than as a quantitative evaluation.

Participants responded positively to the tool's usefulness and readability. They valued having standard metrics immediately available for quick iteration, the worked examples and contextual explanations, and the reporting of job progress; one noted that the evaluation pipeline surfaced metrics they had not previously encountered, and both participants who engaged with the rating scale expressed confidence in the conclusions they drew from it.

Their reservations was mainly on verifiability rather than usability.
Participants could not trace a reported finding back to the individual samples behind it, and one observed that the reports ``take some trust that the tool correctly generated the answers.'' A second was unsure whether the groups shown in the clusterability view were unsupervised clusters or label classes. Participants also asked for pre-specified subgroup analyses and for a scripted path to reproduce an analysis inside a version-controlled repository. Per-participant responses are provided in Appendix~\ref{app:survey}. 
These observations shaped the platform's subsequent development. The current
deployment reports how each metric was computed through a provenance panel,
lets a reported finding be traced back to the individual samples behind it
through per-sample export, distinguishes unsupervised clusters from label
classes in the clusterability view, and adds pre-specified subgroup comparison.
The command-line interface provides the scripted, version-controlled path for
reproducing an analysis that participants requested.

\section{Discussion}
\label{sec:discussion}

We introduced LatentVerse, a framework for unimodal and multimodal representation analysis that unifies diverse metrics for evaluating representation quality within a single report-driven workflow. By combining quantitative diagnostics, visual summaries, comparative
analysis, and descriptive interpretation, LatentVerse supports systematic analysis of latent spaces for users with different levels of technical expertise. This makes representation
analysis useful not only for model evaluation and debugging, but also for discovery-oriented analysis. Our controlled simulations show that LatentVerse helps recover the structural properties of synthetic representations, and our biomedical analyses showed how the same diagnostics characterize what real-world embeddings encode and how the information is embedded. We also conducted a user study to assess the utility of LatentVerse for embedding evaluation in different application domains, and are using the feedback from this study to guide the next iteration of the platform.

The most distinctive contribution of LatentVerse is its support for multimodal representation analysis. As foundation models increasingly combine multiple data modalities, understanding how information is distributed across modalities becomes essential for meaningful validation and trust. LatentVerse addresses this challenge by decomposing fused embeddings into shared and modality-specific components, making it possible to analyze where predictive or biologically relevant signal resides rather than treating the multimodal representation as a single opaque vector. We believe this capability is especially important because it addresses a growing need that is not well served by existing tools and is likely to remain relevant across models, modalities, and application domains.

\subsection{Limitations and future directions}
\label{sec:disc-limitations}
LatentVerse is currently a research prototype and although the platform supports interactive evaluation of latent representations through a web application and command-line interface, further engineering will be needed to support broader deployment across larger user bases, more demanding workloads, and very large embedding datasets. In particular, scaling the platform to support more concurrent users, larger multimodal representations, and more computationally intensive analyses will require additional infrastructure, job scheduling, and resource management.

In terms of utility of LatentVerse in different application domains, our user study currently provides an initial assessment of the platform’s practical value across researchers familiar with machine learning workflows, but it does not yet establish how well the tool transfers across the full range of scientific and application domains in which latent representations are used. Extending this evaluation to additional domains, user profiles, and analytical settings will be a future direction for our work that will also help further develop the application.

\section*{Acknowledgments}
We thank Pia Francesca Rissom, Vít Škrhák, and Zachary Berger for participating in the user study and for their thoughtful feedback, which substantially helped improve the design, usability, and evaluation of LatentVerse.

\noindent S.T.\ is supported by the Eric and Wendy Schmidt Center at the Broad Institute.

\section*{Author Contributions}
M.A.\ developed the LatentVerse platform, including the full-stack web
application and backend, reworked the implementation of the evaluation tests and integrated them into the platform. M.A.\ also handled data curation, helped with the
visualizations, validated the test implementations, ran the experiments,
and led the writing of the paper as well as the conceptualization and
methodology. S.F.\ Helped with running the real-world discovery analysis. J.K.\ helped with the Google Cloud Platform deployment. S.T.\ supervised the research, shaped and guided the methodology for the experiments and contributed to writing the paper. M.M.\ supervised the research and wrote the paper.

\section*{Declaration of Interests}
The authors declare no conflict of interest. 

\section*{Resource Availability}
The LatentVerse evaluation library is openly available at
\url{https://github.com/broadinstitute/ml4h-latentverse}.
The web application is available at \url{https://broad.io/latentverse}. 

UK Biobank data (Section~\ref{sec:biomedical}) are available to approved researchers through the UK Biobank (\url{https://www.ukbiobank.ac.uk}) under application number 7089.

\bibliographystyle{abbrvnat}

\bibliography{ref}

\appendix
\section{Supplemental Information}

\subsection{System architecture and implementation}
\label{app:system}
LatentVerse is organized as a three-tier web application consisting of a
React-based frontend, a Python FastAPI backend, and a standalone machine
learning library. The backend exposes RESTful endpoints that handle file
uploads, input validation, test execution, and result serialization. The
primary endpoint runs the full evaluation pipeline: it locates uploaded
files by request identifier, loads and preprocesses the CSV data, merges
representation and label files on user-specified ID columns, and
dispatches the selected benchmark test.

File ingestion uses a three-tiered strategy keyed on file size. For files
larger than 50~MB, LatentVerse switches to Polars with eight parallel
threads, achieving 10--100$\times$ faster loading compared to pandas for
large tabular files. Files between 10 and 50~MB are handled by PyArrow
with a 1~MB block size and self-destructing buffers to conserve memory.
Smaller files fall back to the pandas C engine.

When multiple label columns are selected for a single test run, the
backend evaluates all columns concurrently using \texttt{asyncio.gather}
combined with a thread pool executor, so wall-clock time scales with the
most expensive label rather than the sum of all labels. To keep this
concurrency from fighting with itself, BLAS thread counts are capped at
one when more than one label is being evaluated in parallel, which avoids
the oversubscription pattern where NumPy and scikit-learn each spin up a
full thread pool inside every asyncio task. Concurrency is also bounded by
a configurable semaphore (\texttt{MAX\_PARALLEL\_LABEL\_TESTS}), and
per-worker thread budgets are capped to prevent BLAS oversubscription.
Disentanglement statistics (entropy, mutual-information gaps) are
accelerated with Numba just-in-time compilation, giving a 10--20$\times$
speedup over the pure-Python path on large representations.

Long-running comparisons run asynchronously. The frontend submits the
job, immediately receives a job ID, and polls a lightweight status
endpoint that reports both a percentage and a named phase
(\texttt{queued}, \texttt{loading\_files}, \texttt{decoupling},
\texttt{running\_tests}, \texttt{generating\_plots}, \texttt{done}). Job
state is tracked in an in-memory store so that a single instance owns each
job for its lifetime, which is also why the deployment pins session
affinity. Multimodal training is off-loaded to a Cloud Run Job via
Pub/Sub, writing inputs and status to GCS and surfacing progress through
the same job pipeline that unimodal tests use, so the frontend treats both
cases identically.

The application is deployed on Google Cloud Run, configured with 8 vCPUs,
16~GiB of memory, autoscaling between zero and ten instances, and CPU
always allocated \texttt{(--no\-/cpu\-/throttling)} so background jobs
continue to make progress between polls. Uploaded files are persisted to a
Google Cloud Storage bucket mounted with GCS-FUSE; completed files are
reaped after 48~hours. For institutional deployments, the platform exposes
an optional authenticated mode with per-user run history and
Identity-Aware Proxy support, currently disabled in the public
deployment. When a compatible GPU is present (NVIDIA L4 on the production
Cloud Run service), MultiLoReFT training runs on-device via PyTorch's CUDA
backend; the service auto-detects hardware at startup and routes
computation accordingly.

\subsection{Web application interface}
\label{app:interface}
The frontend is a single-page application organized around the three-step
workflow described in Section~\ref{sec:workflow}. In the first step, the
user selects an evaluation test; for several tests, contextual messages
explain the label requirements before any files are uploaded. A guidebook
is reachable from the test selector, each comparison card, and the results
panel, providing metric descriptions, score ranges, interpretation
guidance, and common pitfalls. Dark and light themes are supported, a
two-stage reset confirmation prevents accidental loss of a configured
comparison, and an example mode loads pre-computed results for all five
unimodal tests and the multimodal workflow, annotated with interpretive
bullet points.

In the second step, the user configures one or more comparison cards. Each
card accepts an experiment name, a representation file in CSV or TSV
format, and an optional labels file or a small set of comma-separated
numeric labels. After uploading, the application parses the first 100 rows
client-side using PapaParse to auto-detect column headers, identify
headerless files, and flag plausible label columns. A searchable dropdown
lets the user select ID columns for matching representations and labels;
when files are guaranteed to be in the same row order, an injected virtual
column can be used for index-based alignment. An ``advanced'' drawer
exposes run options that affect reproducibility and runtime: a random
seed, an in-place standardization toggle, a row subsample cap, and (for
clustering-based tests) an override for the number of clusters. When
labels are joined on an ID column, the backend tracks how many
representation rows fail to match and surfaces a warning whenever overlap
falls below a configurable threshold. File transfers use the TUS
resumable upload protocol; each file is fingerprinted by name, size, and
last-modified timestamp and is not re-uploaded if an identical fingerprint
is cached and confirmed via a \texttt{/validate-files} check.

In the third step, results are returned as a JSON payload rendered into
four components: an interactive Plotly.js chart whose form depends on the
test (PCA scatter with cluster hulls for clusterability,
mutual-information heatmaps and sorted bar charts for disentanglement,
dimension-removal curves for expressiveness, grouped bar charts with
cross-validation error bars for probing, and noise-degradation line charts
with a 90\%-of-baseline threshold for robustness); a sortable metrics
table; export buttons for raw JSON and formatted CSV; and an optional
AI-generated summary produced by querying the Google Gemini 2.5 Flash via the Vertex AI ``generateContent'' endpoint (service-account / ADC auth).
The Gemini prompt is built server-side from a fixed template that includes
the experiment name, metric definitions with expected ranges,
test-specific interpretation guidance, and the computed values; the model
is instructed to produce a 3--5 sentence technically grounded paragraph
that references the experiment by name and avoids naive good/bad
characterizations. The proxy fronting the Gemini API enforces this system
instruction server-side, strips user-supplied prompt overrides, and
applies a token-bucket rate limit so the summary feature cannot be
repurposed as an open LLM endpoint.

\subsection{Representation evaluation tests}
\label{app:tests}

\subsubsection{Clusterability}
LatentVerse assesses clusterability using both intrinsic and label-aware
metrics. The Silhouette Score \citep{rousseeuw1987silhouettes} and Davies--Bouldin Index
\citep{davies1979cluster} quantify cluster
cohesion and separation directly from the geometry of the representation;
for $N > 5{,}000$, Silhouette is estimated on a 20\% random subsample
(minimum 1,000 samples) for tractability. When ground-truth labels are
available, NMI \citep{strehl2002cluster} measures alignment between discovered clusters and known
classes. We additionally report Cluster Learnability \citep{lu2023using}, the balanced
accuracy of a logistic-regression probe trained on a stratified 80/20
split, which measures how easily class labels can be recovered. In
label-free settings, the analysis relies on the intrinsic metrics alone.

\subsubsection{Disentanglement}
LatentVerse reports the DCI framework \citep{eastwood2018framework} (Disentanglement, Completeness,
Informativeness), the Mutual Information Gap (MIG) \citep{chen2018isolating}, Separated Attribute
Predictability (SAP) \citep{kumar2018variational}, and Total Correlation (TC)
\citep{watanabe1960information}. Formally, for generative
factor $y_j$,
\begin{equation}
\mathrm{MIG}_j = \frac{I(z; y_j)^{(1)} - I(z; y_j)^{(2)}}{H(y_j)},
\end{equation}
where $I(z; y_j)^{(1)}$ and $I(z; y_j)^{(2)}$ are the largest and
second-largest mutual-information values between any single latent
dimension and $y_j$, and $H(y_j)$ is the entropy of the factor. The implementation supports $N \geq 1$ simultaneous generative factors. For
$N > 1$, Disentanglement and Completeness follow the standard DCI weighting \citep{eastwood2018framework}
rather than a plain average: Disentanglement is the per-dimension score (one
minus the normalized entropy of a dimension's relative importance across
factors), averaged over dimensions and weighted by each dimension's total
importance; Completeness is the per-factor score (one minus the normalized
entropy of a factor's relative importance across dimensions), averaged over
factors and weighted by each factor's total informativeness. Informativeness,
MIG, and SAP are averaged over factors with equal weights, and Total
Correlation is computed once for the full representation. Mutual
information is estimated via $k$-nearest-neighbour regression and
classification ($k = 3$) using \texttt{mutual\_info\_regression} and
\texttt{mutual\_info\_classif} from scikit-learn
\citep{pedregosa2011scikit}. TC is approximated via the Gaussian proxy
$\tfrac{1}{2}\log\!\big(\prod_i \sigma_i^2 / \det(\Sigma)\big)$, which is
exact under Gaussian marginals and avoids the cost of GMM-based
estimation.

\subsubsection{Expressiveness}
LatentVerse progressively removes the highest-variance dimensions and
measures the resulting degradation in predictive performance. The removal
schedule sweeps $\{0\%, 10\%, 20\%, 30\%, 40\%, 50\%\}$ of dimensions by
default. Compactness measures the average performance degradation:
\begin{equation}
\mathrm{Compactness} = 1 - \frac{1}{|S|}\sum_{s \in S}\frac{\mathrm{perf}(s)}{\mathrm{perf}(0)},
\end{equation}
where $S = \{0\%, 10\%, 20\%, 30\%, 40\%, 50\%\}$, $\mathrm{perf}(s)$ is the
downstream performance when $s\%$ of dimensions are removed, and
$\mathrm{perf}(0)$ is the full-representation baseline; the score is clamped
to $[0, 1]$. Intrinsic Dimension estimates how many dominant dimensions
can be removed before performance deteriorates substantially.

\subsubsection{Downstream predictability}
LatentVerse uses supervised probes of increasing complexity, from a linear
probe to a five-layer MLP. Each probe is evaluated with three-fold
cross-validation and the mean and standard deviation are reported. Task
type (binary classification, multi-class classification, or regression) is
detected automatically from the label column, and the test reports
Accuracy, AUROC, macro-F1, and $R^2$ as appropriate.\footnote{Exposed as
\texttt{run\_probing} in the underlying LatentVerse library; we use the
more descriptive name throughout the paper.}

\subsubsection{Robustness}
LatentVerse injects Gaussian noise of increasing magnitude into the latent
space at levels $\sigma \in \{0.1, 0.2, 0.3, 0.4, 0.5\}$. In clustering
mode, robustness is reported through Silhouette Score, Davies--Bouldin
Index, and (when labels are available) NMI. In probing mode, it is
evaluated using Accuracy, AUROC, macro-F1, or $R^2$, depending on the
task. For visualization, a representation is considered robust at a given
noise level if its performance remains at or above 90\% of the clean-data
baseline.

\subsection{MultiLoReFT configuration}
\label{app:multiloreft}
Table \ref{tab:multiloreft} summarizes all parameters used for the MultiLoReFT framework to decouple the multimodal information for analysis. 

\begin{table}[h]
\centering
\caption{MultiLoReFT configuration used in our experiments.}
\label{tab:multiloreft}
\begin{tabular}{lc}
\toprule
Hyperparameter & Value \\
\midrule
Shared rank $r_s$ & 4 \\
Modality-specific rank $r_m$ & 4 \\
Learning rate & $1 \times 10^{-3}$ \\
Effective batch size & 1024 \\
Singular-value pruning threshold & 0.1 \\
Restarts (best-of-$N$) & 1 \\
Max epochs / training timeout & 800 / 1200~s \\
\bottomrule
\end{tabular}
\end{table}

\subsection{Simulated datasets for controlled validation}
\label{app:sim}

\subsubsection{Unimodal datasets}
\label{app:sim-unimodal}
Each dataset contains 5,000 samples, 64 real-valued features, and discrete
integer labels, grouped into four families targeting clusterability,
predictability, disentanglement, and expressiveness, with three difficulty
levels each. The clusterability datasets use 5 labels (0--4) and vary how
strongly labels correspond to separable clusters. The predictability
datasets use 10 labels (0--9) and vary the complexity of the feature-to-
label mapping from linearly separable to nonlinear and interaction-driven.
The disentanglement datasets use 10 labels and vary whether label-related
information is encoded in a single axis-aligned factor, spread across
multiple observed dimensions, or mixed across latent factors. The
expressiveness datasets derive 10 labels from a continuous latent score
and vary how compactly predictive information is distributed across the 64
dimensions.

\subsubsection{Multimodal dataset}
\label{app:sim-multimodal}
The simulation generates two modalities from shared and modality-specific
latent variables sampled from non-Gaussian distributions. In total
$n_{\mathrm{hidden}} = 2 + 2 + 2 = 6$ hidden variables are defined: two
shared, two private to modality 1, and two private to modality 2. The
shared variables $z_s \in \mathbb{R}^2$ are sampled as
$z_s \sim \mathrm{Binomial}(1, 0.5) + \mathcal{N}(0, 0.01 I)$.
Modality-specific factors are drawn from distinct non-Gaussian
distributions: $z_{m1} \sim \mathrm{Weibull}(1.5) \times 0.3$ and
$z_{m2} \sim \mathrm{Beta}(3, 2)$. Each point is annotated with four
categorical labels:
\[
y_s \in \{0,1,2,3\},\;
y_{m1} \in \{0,1,2,3\},\;
y_{m2} \in \{0,1,2,3\},\;
y_{\mathrm{joint}} \in \{0,\dots,7\}.
\]
$y_s$ is determined by the unique binary configurations of the un-noised
shared factors. To obtain categorical modality-specific labels, the
private continuous factors are binarized dimension-wise using the
empirical median threshold, yielding a 2-bit codeword per modality;
$y_{m1}$ and $y_{m2}$ are assigned as the index of the corresponding
unique codeword. The joint label $y_{\mathrm{joint}}$ is defined by the
triple $(y_s, y_{m1}, y_{m2})$: each unique triple is mapped to a joint
class ID, deterministically bucketed (via a seed-controlled permutation)
into at most $K = 8$ joint classes. This ensures $y_{\mathrm{joint}}$
depends on all three other labels and is not predictable from either
modality alone.

\subsection{Controlled offline validation supplement}
\label{app:val-supplement}

\begin{figure}[t]
\centering
\includegraphics[width=\textwidth]{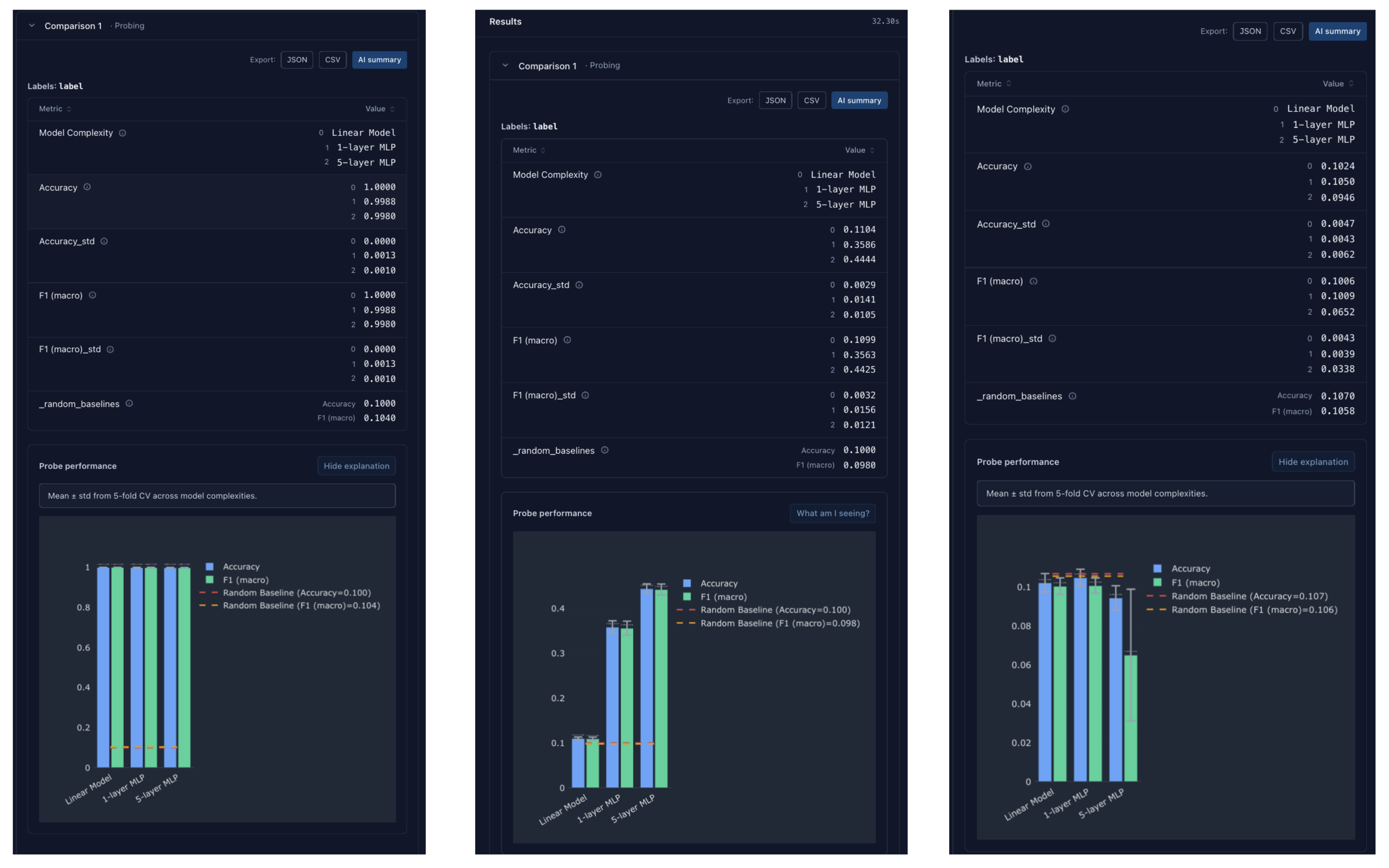}  
\caption{LatentVerse reports for probing-based predictability evaluation
across representations with low, moderate, and high predictive
accessibility. Each report combines a numerical summary of probe
performance across model complexities with a visual comparison of accuracy
and macro-F1 against random baselines.}
\label{fig:probing}
\end{figure}

\paragraph{Predictability.} Across the predictability datasets,
LatentVerse distinguishes directly accessible labels (linear probe
accuracy $= 1.0000$) from labels that require more expressive decoders
(accuracy rising from 0.1104 for the linear probe to 0.4392 for the
5-layer MLP) and from uninformative representations where all probes
remain near random, with a drop in macro-F1 for the most complex probe
indicating instability or overfitting (Figure~\ref{fig:probing}).

\begin{figure}[t]
\centering
\includegraphics[width=\textwidth]{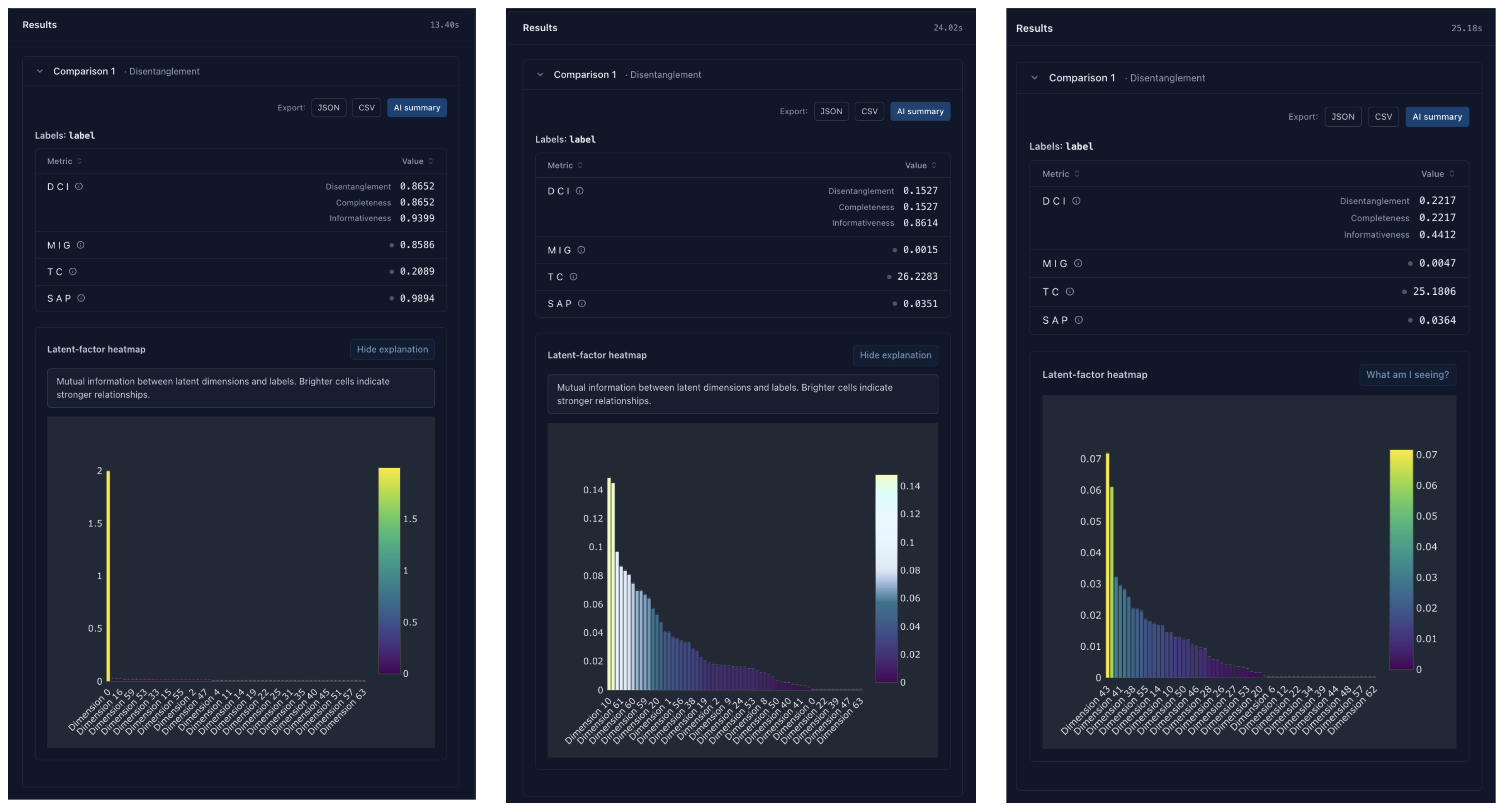}
\caption{LatentVerse reports for disentanglement evaluation across
representations with low, moderate, and high factor entanglement,
combining DCI, MIG, SAP, and Total Correlation with visual summaries of
how label-related information is distributed across latent dimensions.}
\label{fig:disentanglement}
\end{figure}

\paragraph{Disentanglement.} Across the disentanglement datasets, the
easiest setting shows higher DCI, MIG, and SAP with lower Total
Correlation, consistent with signal concentrated in a small number of
dimensions; intermediate and hard settings show weakening scores and
increasing cross-dimension dependence, reflecting progressively more
entangled representations (Figure~\ref{fig:disentanglement}).

\begin{figure}[t]
\centering
\includegraphics[width=\textwidth]{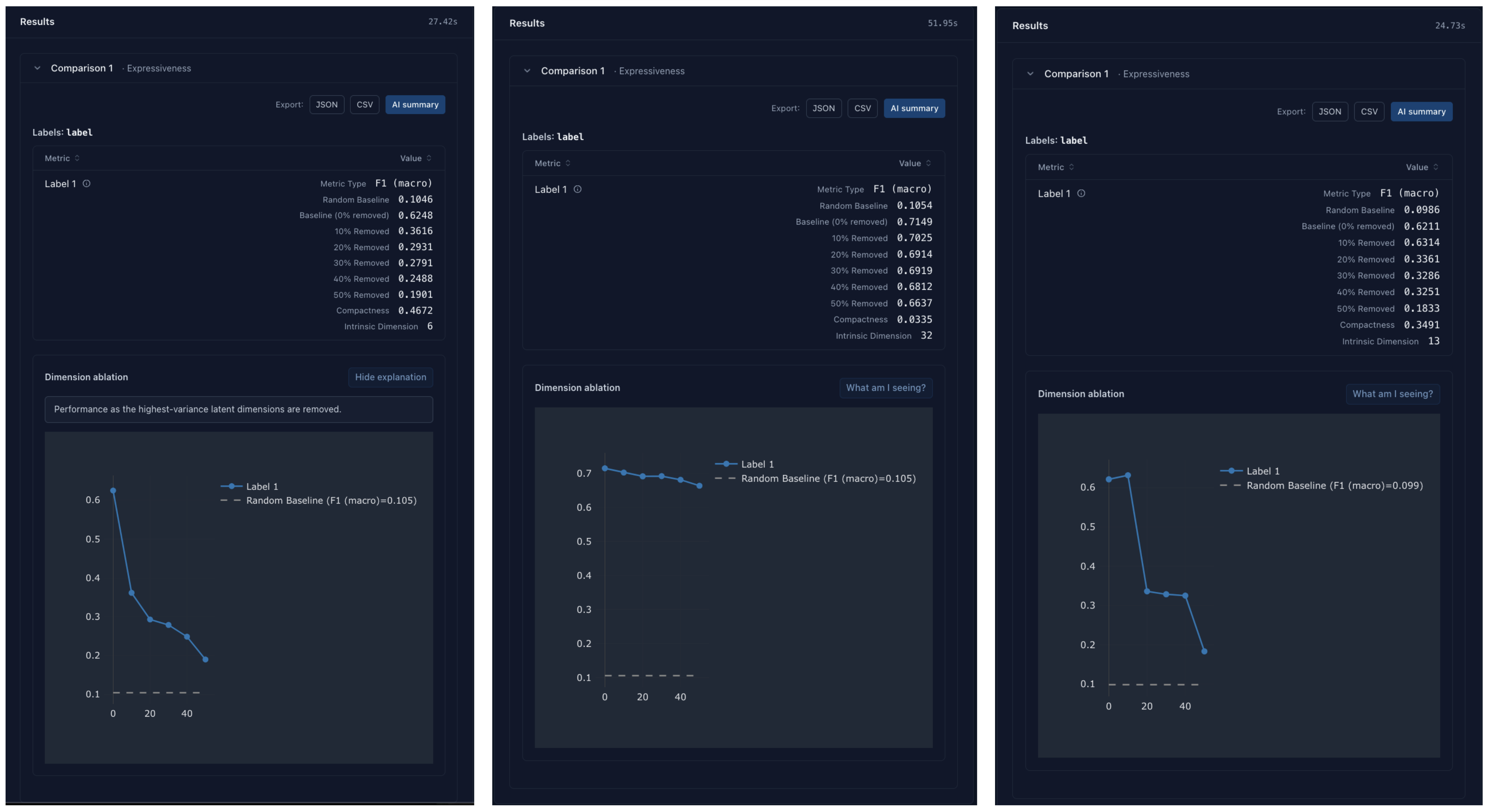}
\caption{LatentVerse reports for expressiveness evaluation across three
synthetic datasets (compact, correlated-redundant, and sparse-informative),
combining baseline performance, compactness, and intrinsic dimension with
an ablation curve showing how predictive performance changes as
high-variance dimensions are progressively removed.}
\label{fig:expressiveness}
\end{figure}

\paragraph{Expressiveness.} In the compact setting, performance drops
sharply under ablation (0.6248 baseline to 0.1901 after removing 50\% of
dominant dimensions; compactness 0.4672, intrinsic dimension 6). In the
semi-compact, correlated setting, performance is largely insensitive to
removal (0.7149 to 0.6637; compactness 0.0335, intrinsic dimension 32). In
the not-compact, sparse setting, performance again drops sharply (0.6211
to 0.1833; compactness 0.3491, intrinsic dimension 13)
(Figure~\ref{fig:expressiveness}).

\subsection{User-study protocol and survey instrument}
\label{app:survey}
To evaluate the usability, interpretability, and perceived utility of
LatentVerse, we designed a short post-use feedback survey, administered
after participants explored the tool using their own latent
representations or embeddings. The survey was designed to understand
(i) what users typically try to evaluate in embeddings, (ii) whether
LatentVerse supports those goals, (iii) whether the tool helps users
interpret and compare representations, and (iv) how it compares with
users' existing workflows.

The survey introduced LatentVerse as a web application for helping developers and researchers analyze and evaluate latent representations through visualizations and quantitative metrics presented in a structured report. Participants were informed that the tool supports comparison, interpretation, and validation of representation properties such as clustering structure, robustness, predictability, expressiveness, and, when relevant, multimodal structure. Participants were also told that responses would be analyzed in aggregate and that anonymized quotes could be used in a research publication.

\subsubsection*{Survey introduction and consent}

Participants first received brief instructions for using the tool. They were told that LatentVerse can be accessed through the web interface and that, in general, the tool requires an embedding file in CSV format, where each row corresponds to a sample, the first column contains a sample identifier, and the remaining columns contain embedding dimensions. Participants were also informed that they could optionally upload a label file with one or more labels per sample, such as class, cohort, site, outcome, or subgroup. These labels support analyses such as separability, subgroup comparisons, and predictability, although some analyses can be run without labels.

Before completing the survey, participants were asked to provide consent for their anonymized responses to be used for research and publication purposes.

\begin{enumerate}
    \item \textbf{Consent.} I consent to my responses being used, in anonymized form, for research/publication purposes.
    \begin{itemize}
        \item Yes, I consent.
    \end{itemize}

    \item \textbf{Optional follow-up.} You may contact me for a short follow-up discussion.
    \begin{itemize}
        \item Yes, you may contact me.
    \end{itemize}

    \item \textbf{Email address.} Participants who agreed to follow-up contact were asked to provide an email address.

    \item \textbf{Recommendation of other users.} Participants were also asked whether they could recommend someone else who may be willing to test LatentVerse.
    \begin{itemize}
        \item Yes, I can recommend someone else.
    \end{itemize}

    \item \textbf{Recommended contact.} Participants who selected this option were asked to provide the recommended person's email address.
\end{enumerate}

\subsection*{About the participant and their workflow}

The first section of the survey collected information about participants' application domains, machine learning workflows, embedding types, and current evaluation practices.

\begin{enumerate}
    \setcounter{enumi}{0}

    \item \textbf{What is your primary application domain?} 
    \begin{itemize}
        \item Clinical / healthcare
        \item Biology / genomics / proteomics
        \item Public health / epidemiology
        \item Neuroscience
        \item General ML / non-biomedical
        \item Other
    \end{itemize}

    \item \textbf{What type(s) of ML are you working with most often?}
    \begin{itemize}
        \item Supervised learning
        \item Self-supervised / contrastive learning
        \item Representation learning / embeddings as a primary output
        \item Generative models
        \item Multimodal models, e.g., text+image or EHR+labs
        \item Time series models
        \item Graph models
        \item LLMs / transformers
        \item Other
    \end{itemize}

    \item \textbf{What do your embeddings most often represent?}
    \begin{itemize}
        \item Patients / individuals
        \item Visits / time windows
        \item Images / patches
        \item Text documents / notes
        \item Cells / genes / proteins
        \item Samples / specimens
        \item Other
    \end{itemize}

    \item \textbf{When in your workflow do you usually evaluate embeddings?}
    \begin{itemize}
        \item Model development / debugging
        \item While choosing between models or checkpoints
        \item Before downstream training
        \item During scientific analysis / hypothesis generation
        \item For reporting in a paper / benchmarking
        \item Teaching / communication
        \item Rarely / never
        \item Other
    \end{itemize}

    \item \textbf{Why do you evaluate embeddings? What decision does it support?} \\
    Free-text response, 1--3 sentences. Participants were given examples such as selecting the best model, understanding what is encoded, checking robustness, comparing checkpoints, verifying biological signal, or detecting subgroup effects.

    \item \textbf{What evaluations do you typically perform on embeddings?}
    \begin{itemize}
        \item Downstream performance, e.g., linear probe, classifier, or regressor
        \item Visualization, e.g., UMAP, t-SNE, PCA, and manual inspection
        \item Clustering and cluster quality metrics
        \item Nearest-neighbor inspection / retrieval sanity checks
        \item Subgroup / cohort comparisons
        \item Outlier / anomaly detection
        \item Robustness checks, e.g., augmentation, noise, or shift
        \item Interpretability / disentanglement analyses
        \item Bias / fairness checks across subgroups
        \item Calibration / uncertainty of embedding-based decisions
        \item Other
    \end{itemize}

    \item \textbf{What tools do you currently use for these evaluations?}
    \begin{itemize}
        \item Python notebooks, e.g., NumPy, PyTorch, or scikit-learn
        \item TensorBoard projector or similar embedding viewers
        \item Custom visualization scripts
        \item Commercial / GUI analytics tools
        \item I do not have a standard approach
        \item Other
    \end{itemize}

    \item \textbf{What usually makes embedding evaluation hard in your setting?} \\
    Participants were asked to select up to two options.
    \begin{itemize}
        \item Too many ad-hoc scripts / poor reproducibility
        \item Hard to compare models consistently
        \item Hard to interpret what metrics mean
        \item Hard to connect results back to specific samples
        \item Takes time to implement validation each time
        \item Unclear what to check / lack of guidance
        \item Hard to reason about multimodal embeddings
        \item Other
    \end{itemize}

    \item \textbf{Can you briefly describe the embeddings you tested today?} \\
    Free-text response. Participants were asked to describe what the embeddings represent, the approximate dataset size, and whether the setting was unimodal or multimodal.
\end{enumerate}

\subsubsection*{Session-specific feedback on LatentVerse}

The second section of the survey asked participants about the analyses they used during their LatentVerse session, the conclusions they reached, and how the tool compared with their usual workflow.

\begin{enumerate}
    \setcounter{enumi}{9}

    \item \textbf{Which analyses or views did you use today?}
    \begin{itemize}
        \item Clusterability
        \item Disentanglement
        \item Expressiveness
        \item Downstream predictability / probing
        \item Robustness
        \item Multimodal shared vs. modality-specific analysis
        \item Visualization of latent space / projections
        \item LLM-generated summary / interpretation
        \item Other
    \end{itemize}

    \item \textbf{How long did it take you to reach a useful conclusion with LatentVerse?}
    \begin{itemize}
        \item $<$ 5 minutes
        \item 5--10 minutes
        \item 10--20 minutes
        \item 20 minutes
        \item I did not reach a useful conclusion
    \end{itemize}

    \item \textbf{Rate the following statements.} \\
    Participants rated each statement on a 1--5 Likert scale, where 1 indicated ``strongly disagree'' and 5 indicated ``strongly agree.''
    \begin{itemize}
        \item LatentVerse helped me answer the evaluation question I cared about.
        \item I could use LatentVerse to compare representations systematically.
        \item I understood what the reported metrics and views were showing.
        \item The tool provided enough evidence to support the conclusions I reached.
        \item I trust the conclusions I reached using LatentVerse.
        \item LatentVerse would be useful in my regular workflow.
    \end{itemize}

    \item \textbf{Compared with your usual workflow, please rate the following.} \\
    Participants rated each statement on a 1--5 Likert scale, where 1 indicated ``strongly disagree'' and 5 indicated ``strongly agree.'' For the multimodal-analysis item, participants were instructed to select neutral if the question was not applicable.
    \begin{itemize}
        \item LatentVerse was faster for this kind of analysis.
        \item LatentVerse made it easier to compare representations.
        \item LatentVerse made the results easier to interpret.
        \item LatentVerse provided clearer evidence than my usual approach.
        \item LatentVerse helped me reach a more confident conclusion.
        \item LatentVerse would be especially useful for multimodal embedding analysis.
    \end{itemize}

    \item \textbf{What conclusion or decision were you able to reach using LatentVerse?} \\
    Free-text response, 1--3 sentences. Participants were given examples such as determining that one model appeared more robust, that a shared representation space was more informative than a modality-specific one, or that an embedding had weak cluster structure.

    \item \textbf{Where did you feel least confident or most confused?} \\
    Free-text response.

    \item \textbf{What did you find most useful about LatentVerse?} \\
    Free-text response.

    \item \textbf{What else would you have liked to do with LatentVerse?} \\
    Free-text response. Participants were prompted to describe what additional information, metric, visualization, or interaction would have made the evaluation more convincing.

    \item \textbf{What is the most important improvement you would like to see?} \\
    Free-text response.

    \item \textbf{Where do you see LatentVerse fitting into your lab, team, or workflow?} \\
    Free-text response.
\end{enumerate}

\subsubsection{Survey Results}

\begin{table}[h]
\centering
\caption{Per-participant responses (1--5). P1 returned the neutral midpoint on
all eight statements and reported evaluating embeddings ``rarely or never.''}
\begin{tabular}{lccc}
\toprule
Statement & P1 & P2 & P3 \\
\midrule
Helps achieve my evaluation goal            & 3 & 4 & 4 \\
I could complete my evaluation              & 3 & 5 & 3 \\
I trust the conclusion I reached            & 3 & 5 & 5 \\
Overall usefulness in my workflow           & 3 & 4 & 4 \\
I understood what each view/metric showed   & 3 & 5 & 4 \\
The UI provided enough context              & 3 & 5 & 4 \\
I could see the evidence behind conclusions & 3 & 3 & 5 \\
I could trace findings back to samples      & 3 & 3 & 3 \\
\midrule
Ease of use (UI/UX)                         & 4 & 4 & 4 \\
\bottomrule
\end{tabular}
\end{table}

\end{document}